\documentclass{article} 
\usepackage{iclr2021_conference,times}

\usepackage{amsmath,amsfonts,bm}

\def\eqref#1{equation~\ref{#1}}

\def\1{\bm{1}}

\DeclareMathAlphabet{\mathsfit}{\encodingdefault}{\sfdefault}{m}{sl}
\SetMathAlphabet{\mathsfit}{bold}{\encodingdefault}{\sfdefault}{bx}{n}

\usepackage{hyperref}
\usepackage{url}

\title{Parameter-tuning-free data entry error \\ unlearning with adaptive selective \\ synaptic dampening}

\author{Stefan Schoepf\textsuperscript{\rm 1}, Jack Foster\textsuperscript{\rm 1,2} \& Alexandra Brintrup\textsuperscript{\rm 1,2}\\
\textsuperscript{\rm 1} University of Cambridge, Department of Engineering\\
\textsuperscript{\rm 2} The Alan Turing Institute\\
\texttt{\{ss2823,jwf40,ab702\}@cam.ac.uk} \\
}

\usepackage{algorithm}
\usepackage{algorithmic}
\usepackage{amssymb}
\usepackage{graphicx}
\usepackage{subcaption}
\usepackage{multirow}

\usepackage{tcolorbox}

\iclrfinalcopy 
\begin{document}

\maketitle

\begin{abstract}

Data entry constitutes a fundamental component of the machine learning pipeline, yet it frequently results in the introduction of labelling errors. When a model has been trained on a dataset containing such errors its performance is reduced. This leads to the challenge of efficiently unlearning the influence of the erroneous data to improve the model performance without needing to completely retrain the model. While model editing methods exist for cases in which the correct label for a wrong entry is known, we focus on the case of data entry errors where we do not know the correct labels for the erroneous data. Our contribution is twofold. First, we introduce an extension to the selective synaptic dampening unlearning method that removes the need for parameter tuning, making unlearning accessible to practitioners. We demonstrate the performance of this extension, adaptive selective synaptic dampening (ASSD), on various ResNet18 and Vision Transformer unlearning tasks. Second, we demonstrate the performance of ASSD in a supply chain delay prediction problem with labelling errors using real-world data where we randomly introduce various levels of labelling errors. The application of this approach is particularly compelling in industrial settings, such as supply chain management, where a significant portion of data entry occurs manually through Excel sheets, rendering it error-prone. ASSD shows strong performance on general unlearning benchmarks and on the error correction problem where it outperforms fine-tuning for error correction. 
\end{abstract}
\textbf{Competing interests:} No competing interests

\section{Introduction}


Data is essential for training machine learning models but as \citet{northcutt2021pervasive} investigated, even the ten most widely used datasets in the literature contain an average of 3.3\% label errors. When errors are detected after a model has been trained, there are three options: Continue using a model that makes predictions based on partially false data, retrain the model from scratch, or attempt to fix the model post-hoc. Fully retraining and redeploying a new model is not only time-intensive but also potentially costly with large models that use significant compute resources for training. Fixing an already deployed model is therefore often the preferable option.

\begin{figure}[]
    \centering
    \includegraphics[width=1\textwidth]{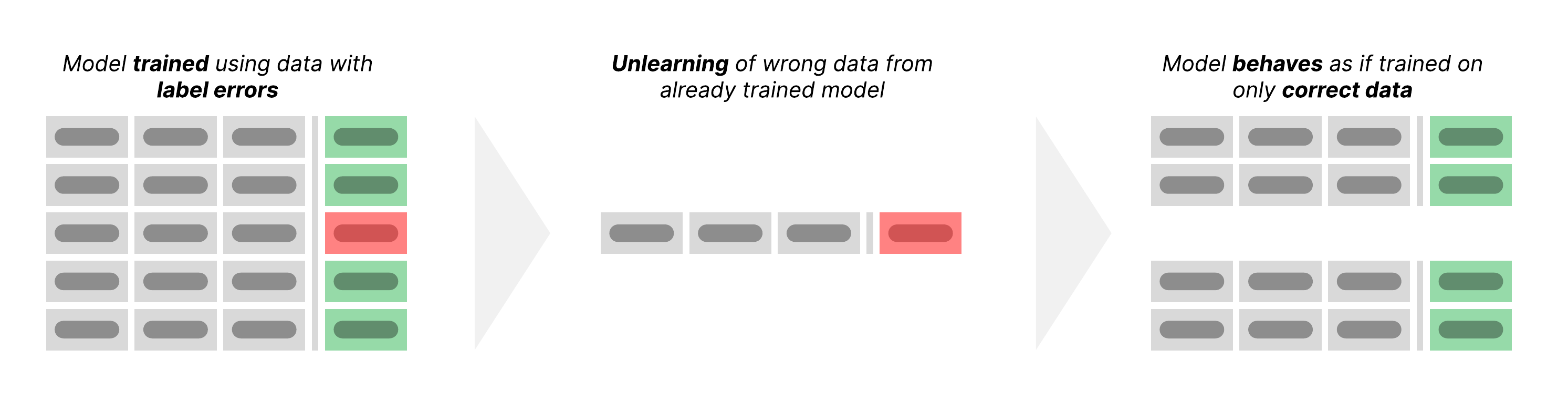}
    \caption{Unlearning data entry errors from a model trained on data including label errors can achieve model behaviour similar to that of a model trained solely on the correct data.}
    \label{fig:illust}
\end{figure}

Fixing a model can fall into two categories. When we know the correct labels, model editing approaches can be used to re-guide the wrongly labelled observations to the correct label (e.g., for LLMs \citep{edit_hartvigsen2023aging}, classifiers \citep{edit_Wang_2022},  generalized additive models \citep{edit_santurkar2021editing}). If the correct labels are unknown, as is often the case in practice, we want to remove the influence of these wrongly labelled observations from the model to make it behave as if it was only trained on the correctly labelled data as shown in Fig. \ref{fig:illust}. Removing data from an already trained model, referred to as machine unlearning, is a non-trivial problem that aims to balance retaining the accuracy of the model while removing the influence of the to-be-forgotten data from the model \citep{cao2015towards}. Existing methods have two shortcomings for the task of data entry error forgetting. First, multiple state-of-the-art methods use 'confusion' approaches in which they assign random or deliberately wrong labels to the observations that should be forgotten to make the model forget the observation \citep{chundawat2023can,tarun2023fast,kurmanji2023unbounded, kodge2023deep}. In our case, this is undesirable, as this can lead to further wrong predictions given that the features are a real occurrence, only the labels are wrong. \citet{kodge2023deep} shows the effect of their unlearning method in a confusion matrix before and after forgetting where in a 10 class forgetting task after forgetting the class cat, cats go from a 0.3\% chance of being classified as a ship to 11.4\% (versus 26.2\% chance to be classified as a dog). Cats are forgotten but the behaviour of the model when encountering a cat seems highly undesirable. A state-of-the-art method that does not suffer from this limitation is Selective Synaptic Dampening (SSD) \citep{foster2023fast}. But SSD shares the second shortcoming with the other state-of-the-art methods, a reliance on correctly chosen parameters to balance the aggressiveness of forgetting and the retaining of model performance.

We address this shortcoming with an extension to SSD that automatically chooses a suitable aggressiveness/performance trade-off. This enables practitioners to use unlearning for data entry error unlearning out of the box without any parameter tuning.

We first show the performance of Adaptive Selective Synaptic Dampening (ASSD) on unlearning benchmarks using Resnet18 (RN) and Vision Transformer (ViT) before applying ASSD to a data entry error unlearning scenario using real-life delivery delay data from an e-commerce company \citep{dataset} where we compare ASSD performance to retraining and fine-tuning. The model parameter counts used across these tasks range from 4.500 (simple neural network) to 85.000.000 (ViT) combined with unlearning shares of $<$1\%-10\% of the full dataset. This large spread showcases the robustness of our adaptive method.

We make the following key contributions:


\begin{enumerate}
    \item We develop an adaptive parameter selection method for the method of \citet{foster2023fast} to enable parameter-tuning-free usage of unlearning in practice.
    \item We demonstrate the performance of ASSD to unlearn data entry errors on a supply chain delay prediction problem using real-life data.
\end{enumerate}

\section{Related Work}
\label{sec:related}

\citet{tanno2022repairing} have laid the groundwork for the problem of identifying training samples that cause model failures (e.g., mislabeled data) and then unlearning these datapoints to improve the model. 


The framework of \citet{tanno2022repairing} uses a Bayesian view of continual learning and Bayesian unlearning. Since then, new state-of-the-art methods in machine unlearning have been added to the literature. Recent high-performing methods include \citet{chundawat2023can}, \citet{tarun2023fast}, \citet{kurmanji2023unbounded}, and \citet{kodge2023deep}. \citet{kurmanji2023machine_database} uses, amongst other worse-performing methods, the method from \citet{kurmanji2023unbounded} to unlearn outdated information from models as database data changes. The mentioned methods have in common, that they aim to degrade the performance of the model on $D_f$ using additional training epochs (e.g., by assigning purposefully wrong labels to $D_f$) while retaining performance on $D_r$ to achieve unlearning. 
In the task of wanting to unlearn mislabeled data, this can possibly result in a critical problem. Given a supply chain example, we might have an order that \textit{arrived early} but was mislabeled as \textit{arriving on time}. If an unlearning method aims to degrade the performance of $D_r$, to which this example belongs, and makes it predict \textit{late delivery}, the resulting model would behave in an undesirable way by mislabeling such an observation again, this time to another wrong label. 
\citet{kodge2023deep} shows this undesirable behaviour of their unlearning method in a confusion matrix. After unlearning the class cat in a ten-class dataset, cats go from a 0.3\% chance of being classified as a ship to 11.4\% even though a much more similar class, namely dogs, exists.
A method that does not suffer from this limitation while being competitive with the performance of the aforementioned methods is Selective Synaptic Dampening (SSD) from \citet{foster2023fast}. SSD builds upon the work of \citet{golatkar2020eternal} (Fisher forgetting) that uses the Fisher Information Matrix (FIM) to determine how important each model parameter is to $D_r$ and $D_f$. SSD achieves significant improvements compared to Fisher forgetting, making it competitive with the aforementioned state-of-the-art methods both in terms of speed as well as performance.

A further limitation that the mentioned unlearning methods share is the reliance on correctly chosen parameters to balance the "aggressiveness" of unlearning with the retention of performance on $D_r$. In the case of SSD, these are the parameters $\alpha$ and $\lambda$. \citet{foster2024zeroshot} shows significant performance dips of SSD and other methods such as SCRUB from \citet{kurmanji2023unbounded} when used without an extensive parameter search per unlearning task.

Our method addresses this limitation of SSD to make it parameter-free and easy to use in practice.

\section{Preliminaries}

Analogous to \citet{foster2023fast}, the full dataset $\mathcal{D}=\{x_{i}, y_{i}\}^{N}_{i=1}$ with training samples $x_{i}$ and class labels  $y_{i} \in \{1,...,K\}$ acts as the starting point on which we train a baseline model. We then aim to remove the influence of the forget set $\mathcal{D}_{f} \subset \mathcal{D}$ from the baseline model to make it behave as if it were trained on the retain set only $\mathcal{D}_{r} = \mathcal{D} \setminus \mathcal{D}_{f} $. $D_r$ for an error correction scenario denotes the subset of the data without errors while $D_f$ represents the data with label errors. The difficulty hereby is keeping the model performance (e.g., accuracy) high in regards to $D_r$ while forgetting the information the model learned from $D_f$. The model parameters are referred to as $\theta \in \mathbb{R}^{m}$.

\citet{foster2023fast} calculate the model importances using FIM before modifiying the baseline model via the selection of overly specialised parameters in Eq. \ref{eq:ssd1} and then dampening these parameters by $\beta$ as shown in Eq. \ref{eq:ssd2}. The first requires the selection of an appropriate parameter $\alpha$, the latter $\lambda$. Choosing unsuitable parameters either leads to a lack of forgetting or overly aggressive forgetting leading to a drastic drop in model performance.

\begin{equation}    
    \theta_{i} = 
        \begin{cases}    
            \beta\theta_{i}, & \text{if } []_{\mathcal{D}_{f,i}} > \alpha[]_{\mathcal{D},i}\\
            \theta_{i}, & \text{if } []_{\mathcal{D}_{f,i}} \leq \alpha[]_{\mathcal{D},i}  
    \end{cases}\quad
    \forall i \in [0,|\theta|]\quad\quad
\label{eq:ssd1}
\end{equation} \\

\begin{equation}    
    \text{where } \beta = min(\frac{\lambda []_{\mathcal{D},i}}{[]_{\mathcal{D}_{f,i}}}, 1)
\label{eq:ssd2}
\end{equation}

\section{Proposed Method}
\label{sec:method}

Due to the sheer number of possible model architectures, training methods, data types, etc., a universal theoretical guarantee for selecting good SSD parameters to achieve unlearning is practically intractable. Therefore, we create a general parameter selection approach building upon three key assumptions and validating the performance experimentally across a range of tasks, models, and datasets. The key assumptions for ASSD are:

\begin{tcolorbox}
    \bf{Assumption 1:}\\
    \normalfont{
        The FIM-based selection and dampening process of SSD can be simplified to only rely on $\alpha$, keeping $\lambda=1$ and thus reducing the problem complexity significantly while not harming performance.
    }
\end{tcolorbox}

\begin{tcolorbox}
    \bf{Assumption 2:}\\
    \normalfont{
        Class frequencies and subpopulation frequencies within classes have a long tail and follow a power law, closely resembling a log-log distribution as described by \citet{shortlongtail}.
    }
\end{tcolorbox}

\begin{tcolorbox}
    \bf{Assumption 3:}\\
    \normalfont{
        With an increasing amount of data to be unlearned, overlaps in FIM parameter importances occur and thus allow the removal of data with a sub-linearly scaling increase in parameters to be modified.
    }
\end{tcolorbox}

\begin{figure}[]
    \centering
    \subfloat[\centering Parameter influence on retain data accuracy]{{\includegraphics[width=6.5cm]{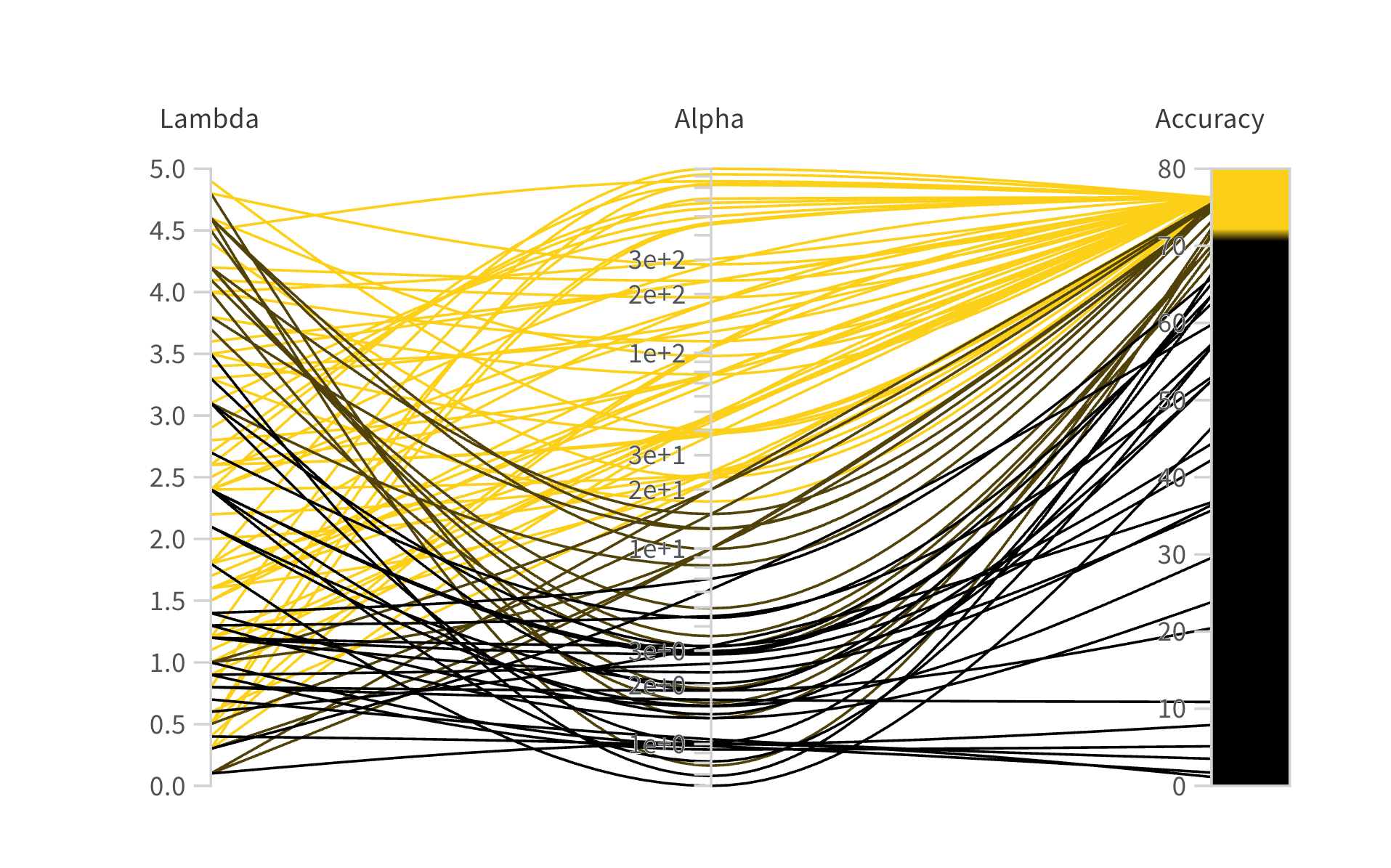} }}%
    \qquad
    \subfloat[\centering Parameter influence on unlearning performance]{{\includegraphics[width=6.5cm]{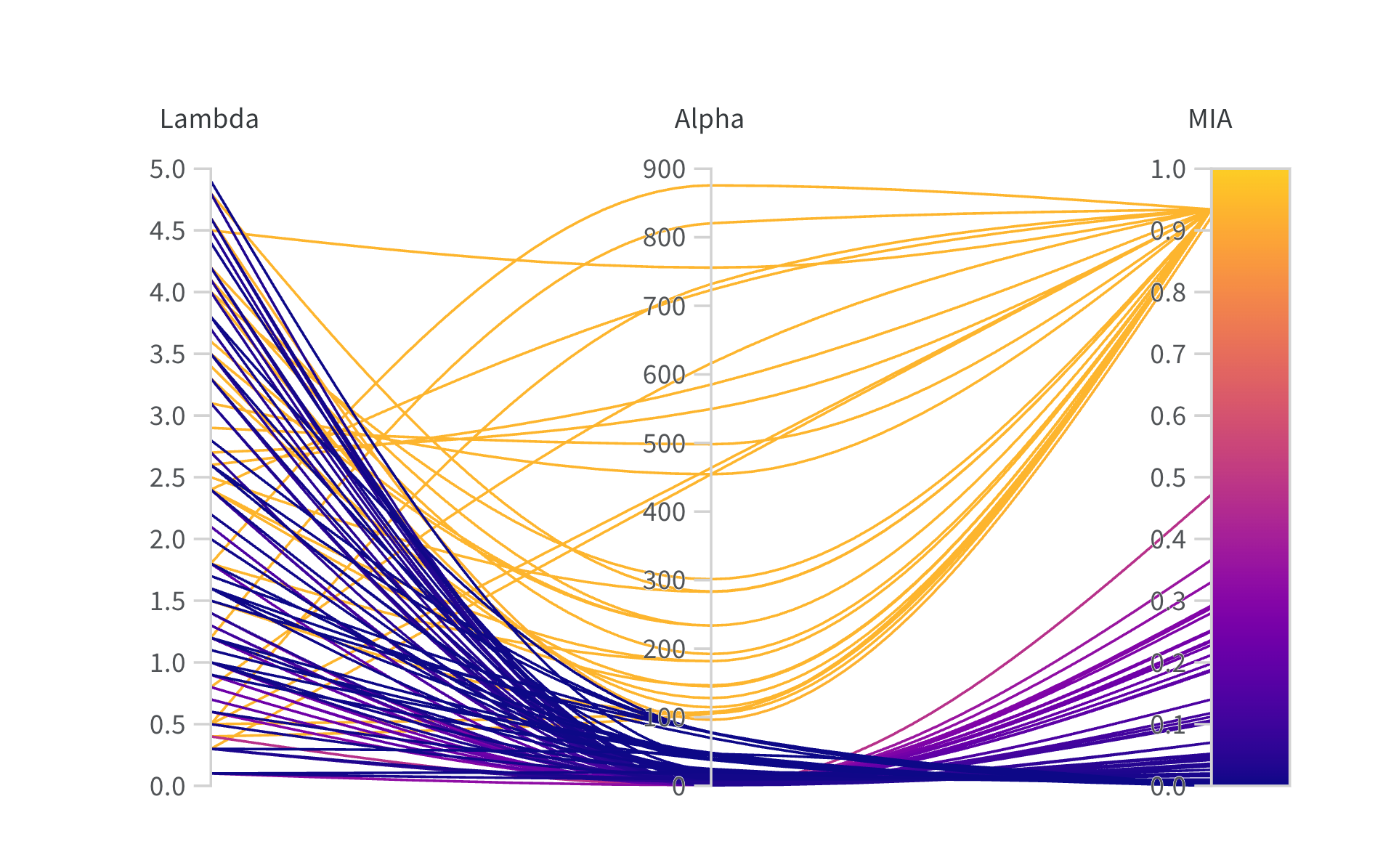} }}%
    \caption{SSD hyperparameter search on the CIFAR100 rocket fullclass unlearning task on ResNet18 using Optuna highlights the high importance of $\alpha$ whereas $\lambda$ values show little impact on the unlearning performance as quantified by the MIA and retain accuracy values.}%
    \label{fig:sensitivity}%
\end{figure}

Assumption 1 acts as the foundation for ASSD by simplifying the problem to a single parameter. Our assumption that the importance of $\alpha$ far outweighs $\lambda$ to the degree that the latter can be kept constant without performance deterioration of SSD is supported by the results shown in Fig. \ref{fig:sensitivity}. We observe that the magnitude of $\lambda$ does not have a notable influence on either the used Membership Inference Attack (MIA), which is used to assess the quality of unlearning, or the accuracy of the unlearned model. The only notable exception is $\lambda=0$, where we can observe a significant drop in model accuracy. This aligns with the method motivation of \citet{foster2023fast} that dampening leads to better performance than simply setting a parameter to zero which can then act like a dead neuron destroying model performance. Due to this behaviour, we set $\lambda=1$ for ASSD which simplifies equation \ref{eq:ssd2} to equation \ref{eq:beta}.

\begin{equation}   
     \text{where } \beta = min(\frac{[]_{\mathcal{D},i}}{[]_{\mathcal{D}_{f,i}}}, 1)
\label{eq:beta}
\end{equation} \\

\begin{figure}[]
    \centering
    \includegraphics*[width=0.6\columnwidth]{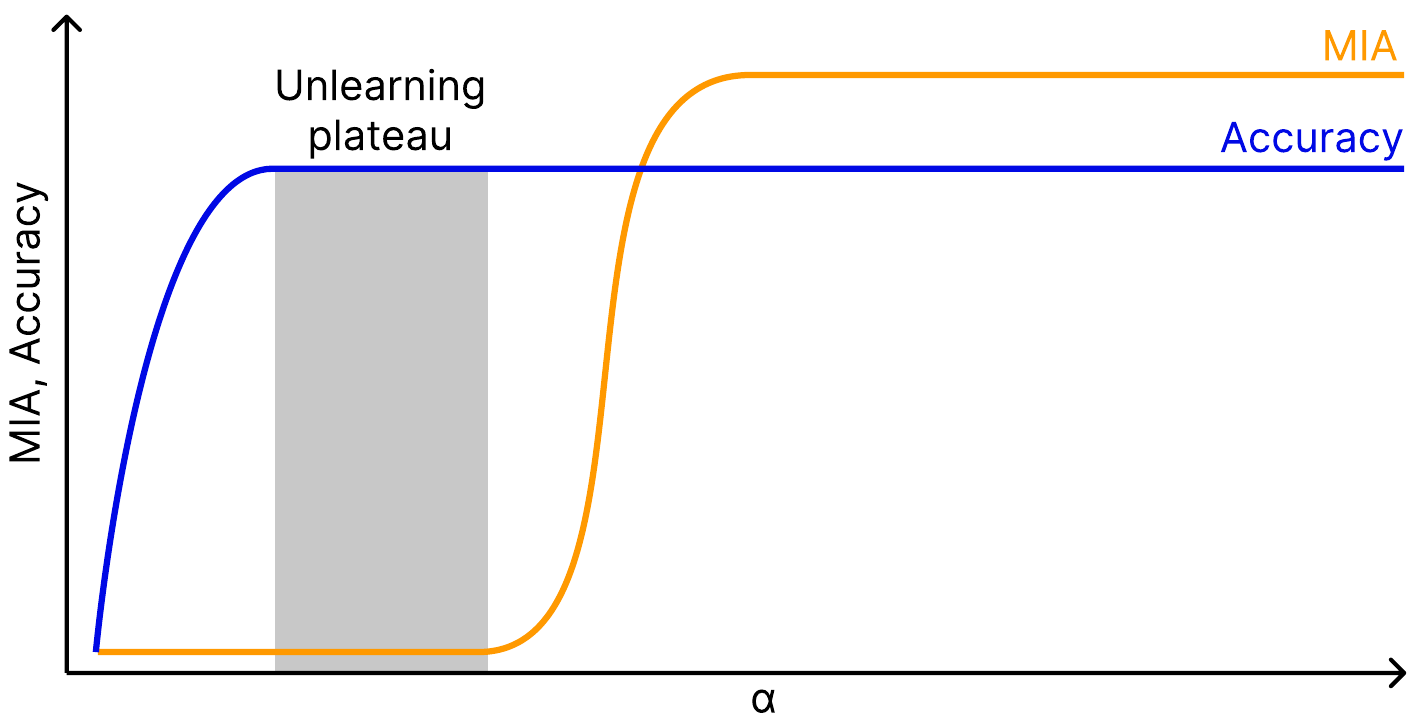}
    \caption{A simplified visualization of the unlearning plateau where model accuracy and MIA both are at near ideal values. Setting $\alpha$ above or below the plateau leads to significant drops in the desired performance of MIA or accuracy.}
    \label{fig:plateau}
\end{figure}

With equation \ref{eq:beta} we observe the behaviour shown in Fig. \ref{fig:plateau}. For each unlearning task (e.g., unlearning the CIFAR100 class rocket from a ResNet18 model), we assume there exists a plateau where MIA and retain accuracy are at a near-optimal level with some fluctuation along the plateau. The accuracy plateau can be explained by the fact that with a sufficiently high $\alpha$ no or minimal changes to the model are performed and thus no performance degradation takes place. Equally, no or minimal changes lead to a stable MIA while overly aggressive unlearning also leads to a low MIA but at the cost of accuracy (i.e., model loses all knowledge). The location and width of this plateau changes not only with each task type (e.g., fullclass versus sub-class unlearning) but also when unlearning different (sub)classes. When unlearning rockets from CIFAR100, the MIA starts to increase after $\alpha=80$ while unlearning the rocket sub-class from CIFAR20 leads to a MIA increase starting at $\alpha=25$. In contrast, the accuracy plateau is reached at around $\alpha=10$ for CIFAR100 rocket full-class and $\alpha=5$ for CIFAR20 rocket sub-class. Even though we unlearn the same images in both cases, the plateau is much smaller in the sub-class task (5-25) than in the full-class task (10-80). We can even observe a shift of the plateau in the original method paper of \citet{foster2023fast} where they report that the class baby observes a critical drop in performance when using the same $\alpha$ as the other classes for the fullclass forgetting task. Due to this shifting nature of the plateau from task to task, we need an adaptive approach for use in practice.

Our approach determines a viable $\alpha$ parameter using assumptions 2 and 3. 
\citet{shortlongtail}, shows that for most datasets class frequencies and subpopulation frequencies within classes have a long tail and follow a power law, closely resembling a log-log distribution. We further assume that forgetting multiple similar images in such a population at once can be achieved with fewer parameters to be changed than the sum of each of them individually (e.g., forgetting all rockets out of a dataset vs two specific rockets). This means that with an increasing number of samples to be forgotten, there will be some kind of overlap that can be used for more efficient forgetting, thus not requiring as many parameters as forgetting each sample individually. Putting our log-log distribution assumption together with shared memorization parameters between similar images, we propose a two-step selection process for $\alpha$.

Our method selects the x-th percentile of relative importance values as the ones that are suspected to contain overly specialized information, i.e. memorization of the forget samples. This is achieved by combining the log assumption from assumption 1 with the unlearning synergy assumption from assumption 2 in equation \ref{eq:percentile}. Equation \ref{algo:assd} then uses the calculated percentile $p$ to determine $\alpha$ from the distribution of relative FIM importances. This adaptive parameter selection step is easily integrated in the SSD workflow as shown in Algorithm \ref{algo:assd}. The calculations of $p$ and $\alpha$ take part after $[]D_f$ and only adds the computational complexity of sorting the relative importances in equation \ref{eq:percentile} for percentile selection which is far lower than the compute time complexity of $[]D_f$ and $[]D$. As with the original SSD, the computationally expensive $[]D$ only needs to be calculated once and can then be stored.

\begin{equation}   
     \text{where } \alpha = P_p(\frac{[]_{\mathcal{D}}}{[]_{\mathcal{D}_{f}}}), \quad
     p \in [0,100]\quad\quad
\label{eq:percentile}
\end{equation} \\

\begin{equation}   
     \text{where } p = 100 - log \left( 1 + \frac{|D_f|}{|D|}\cdot 100 \right)
\label{eq:percentile}
\end{equation} \\


\begin{algorithm}[]
\caption{Adaptive Selective Synaptic Dampening}
\label{alg:algorithm}
\textbf{Input}: $\phi_{\theta}$, $\mathcal{D}$, $\mathcal{D}_f$\\
\textbf{Parameters}: None\\
\textbf{Output}: $\phi_{\theta'}$
\begin{algorithmic}[1] 
\STATE Calculate and store $[]_{\mathcal{D}}$ once. Discard $\mathcal{D}$.
\STATE Calculate $[]_{\mathcal{D}_f}$
\STATE Calculate $p$
\STATE Calculate $\alpha$
\FOR{i in range $|\theta|$}
\IF {$[]_{\mathcal{D}_{f}, i} > \alpha []_{\mathcal{D}, i}$}
\STATE  $\theta'_{i} = min(\frac{[]_{\mathcal{D},i}}{[]_{\mathcal{D}_{f,i}}}\theta_{i}, \theta_{i}) $
\ENDIF
\ENDFOR
\STATE \textbf{return} $\phi_{\theta'}$
\end{algorithmic}
\label{algo:assd}
\end{algorithm}

\section{Experimental Setup}

We perform two types of experiments to empirically validate the performance of ASSD. First, we benchmark ASSD against SSD to show that the automatic parameter selection works across different unlearning tasks and model types. Second, we show the applicability of ASSD for error unlearning on a real-life tabular supply chain dataset.

\subsection{Datasets}
For the ASSD versus SSD benchmarking, we use the same benchmarks as \citet{foster2023fast} and \cite{chundawat2023can}. These are the image classification datasets CIFAR10, CIFAR20, CIFAR100 \citep{CIFAR_krizhevsky2010convolutional} and PinsFaceRecognition \cite{pinsds}.

For the error unlearning task, we use the supply chain dataset of \citet{dataset} which includes clothing, sports, and electronic part data from an e-commerce company including delivery delay that we use as labels (i.e., early, on-time, delayed). 

\subsection{Models}
For the unlearning benchmarks, we use ResNet18 \citep{he2016deep} and Vision Transformer \citep{dosovitskiy2020image_vit} as done in \citet{chundawat2023can} and \citet{foster2023fast}. The training of the models is done as described in \citet{foster2023fast}

For the error unlearning task we use three different sizes of simple fully connected neural networks to also showcase the performance of ASSD on much smaller models with 4.500+ parameters versus the 85M of ViT. The three models are two layers of 50 neurons, three layers of 100 neurons, and five layers of 250 neurons to show that results hold up across different network sizes. We train the models with batch normalization \citep{ioffe2015batch}, ReLu activation \citep{agarap2018deep} and stochastic gradient descent \citep{amari1993backpropagation}. The epochs are set to 25 with a learning rate decay of 0.95 and Optuna \citep{akiba2019optuna} is used to find an optimal learning rate and regularization on the full dataset without any labelling errors. These parameters are then used to train the models that contain label errors.

All models are trained on an NVIDIA RTX4090 with Intel Xeon processors.

\subsection{Unlearning tasks}

For the unlearning benchmarking task of ASSD versus SSD, we use the same three unlearning scenarios as used by \citet{foster2023fast} and \citet{chundawat2023can}: (i) Single-class forgetting, (ii) sub-class forgetting, and (iii) random observations forgetting. In (i) we perform unlearning on CIFAR100, CIFAR 20, and PinsFaceRecognition. We unlearn full classes of each (e.g., rockets out of CIFAR100, the face of a person out of PinsFaceRecognition). In (ii) we forget a subclass from CIFAR20. CIFAR20 contains the same classes of CIFAR100 grouped into 20 superclasses that share the same label in CIFAR20 (e.g., the rocket subclass gets labelled as part of vehicles). In (iii) we use CIFAR10 and select samples randomly across all classes to be forgotten. Due to the high similarity of CIFAR10 samples in the classes, unlearning in this scenario is very minor as forgetting one image of an aeroplane highly likely still leads to the correct classification of this image as an aeroplane due to the numerous other remaining images of similar looking aeroplanes in the dataset.

Our error unlearning task can also be seen as (iii) random unlearning. We use the e-commerce dataset of \citet{dataset} to predict the expected delivery time of an order (early, on-time, delayed). Depending on the error amount we select a percentage of $D$ that we randomly mislabel (e.g., on-time to delayed). The mislabeled data $D_f$ is then unlearned to improve model performance.

\subsection{Evaluation and baselines}

We use two different ways to evaluate the performance of ASSD. In the general unlearning benchmark, we compare ASSD against SSD, and model retraining (i.e., training the model again on only $D_r$). The baseline values reported are from the original model trained on 
$D$. Evaluation is performed using the model accuracy on $D_r$ and $D_f$ as well as the forgetting performance using a Membership Inference Attack (MIA). MIA aims to determine if a sample of the dataset was in the training data or not. We use the logistic regression MIA implementation used in \citet{chundawat2023can} and \citet{foster2023fast} for comparability. While $D_r$ should remain as high as possible to retain performance on the remaining data, $D_f$ and MIA are more difficult to evaluate. As described in \citet{chundawat2023can}, an unlearned model that behaves similarly to a retrained model that has never seen the to-be-forgotten data would be ideal. Due to generalization (e.g., other cars in a dataset allow the model to correctly classify a forgotten car as a car) $D_f$ and MIA do not necessarily need to be zero or close to zero to behave as desired. \citet{chundawat2023can} even highlights the danger of the \textit{Streisand effect} when lowering $D_f$ or MIA far below what a retrained model would do. 

In the error unlearning task we report the model accuracy of a baseline model trained on $D$ including the label errors, a retrained model only trained on $D_r$, a fine-tuned model (i.e., baseline model with one additional fine-tuning epoch on $D_r$), and ASSD. Unlike the unlearning tasks, the performance on future unseen data is essential for this application. Therefore, we exclude the last 20\% of the full dataset (by timestamps of the data) from the baseline training data. $D$ denotes the training data with $D_r\subset D$ and $D_f = D \backslash D_r$. A higher accuracy on $D_r$ (train data excluding errors) and the test data is desired. As retraining is significantly more time intensive, the comparison focuses on fine-tuning for real-life application. We use one fine-tuning epoch with a learning rate further decreased beyond the learning rate of the last training epoch by the learning rate decay parameter $\gamma=0.95$. This resembles a simple yet effective approach in practice and as shown in the results section, and does not overfit on the train data.

\section{Results and Discussion}

We first compare ASSD and SSD on the unlearning benchmarks used in \cite{foster2023fast} and \cite{chundawat2023can} before applying ASSD to an error unlearning task on e-commerce delay prediction. The first part focuses on the ability to remove information from a model so that an attacker cannot determine if certain data was used in the training process (e.g., data privacy applications) while preserving model performance. The second part focuses on the ability to remove erroneous data from a model that can lead to wrong predictions. The focus hereby is on improved model prediction performance.

\begin{table}[]
\centering
\fontsize{8pt}{10pt}\selectfont
\setlength{\tabcolsep}{1.7pt}
\begin{tabular}{ll|cccc|cccc}
                            &                   & \multicolumn{4}{c|}{ResNet18}                                                                                                                                                   & \multicolumn{4}{c}{Vision Transformer}                                                                                        \\ \hline
\multicolumn{1}{l|}{Class}  & metric            & \multicolumn{1}{c|}{baseline}       & \multicolumn{1}{c|}{retrain}                                   & \multicolumn{1}{c|}{ASSD}      & SSD                                     & \multicolumn{1}{c|}{baseline}       & \multicolumn{1}{c|}{retrain}        & \multicolumn{1}{c|}{ASSD}  & SSD                  \\ \hline
\multicolumn{1}{l|}{baby}   & $\mathcal{D}_{r}$ & \multicolumn{1}{c|}{76.38$\pm$0.00} & \multicolumn{1}{c|}{73.10$\pm$0.55}                            & \multicolumn{1}{c|}{76.10$\pm$0.00}     & \textbf{44.05$\pm$0.00}                         & \multicolumn{1}{c|}{88.93$\pm$0.00} & \multicolumn{1}{c|}{90.27$\pm$0.15} & \multicolumn{1}{c|}{88.31$\pm$0.00} & 88.59$\pm$0.00       \\
\multicolumn{1}{l|}{}       & $\mathcal{D}_{f}$ & \multicolumn{1}{c|}{72.48$\pm$0.00} & \multicolumn{1}{c|}{0.00$\pm$0.00}                             & \multicolumn{1}{c|}{0$\pm$0.00}         & 0.00$\pm$0.00                           & \multicolumn{1}{c|}{90.19$\pm$0.00} & \multicolumn{1}{c|}{0.00$\pm$0.00}  & \multicolumn{1}{c|}{0$\pm$0.00}     & 0.00$\pm$0.00        \\
\multicolumn{1}{l|}{}       & MIA               & \multicolumn{1}{c|}{92.60$\pm$0.00} & \multicolumn{1}{c|}{2.44$\pm$0.01}                             & \multicolumn{1}{c|}{0.2$\pm$0.00}       & 5.40$\pm$0.00                          & \multicolumn{1}{c|}{75.60$\pm$0.00} & \multicolumn{1}{c|}{21.53$\pm$0.03} & \multicolumn{1}{c|}{2.4$\pm$0.00}   & 0.60$\pm$0.00        \\ \hline
\multicolumn{1}{l|}{lamp}   & $\mathcal{D}_{r}$ & \multicolumn{1}{c|}{76.39$\pm$0.00} & \multicolumn{1}{c|}{72.89$\pm$0.34}                            & \multicolumn{1}{c|}{76.36$\pm$0.00}     & 76.08$\pm$0.00                          & \multicolumn{1}{c|}{88.84$\pm$0.00} & \multicolumn{1}{c|}{90.10$\pm$0.19} & \multicolumn{1}{c|}{88.64$\pm$0.00} & 89.06$\pm$0.00       \\
\multicolumn{1}{l|}{}       & $\mathcal{D}_{f}$ & \multicolumn{1}{c|}{70.49$\pm$0.00} & \multicolumn{1}{c|}{0.00$\pm$0.00}                             & \multicolumn{1}{c|}{0$\pm$0.00}         & 0.00$\pm$0.00                           & \multicolumn{1}{c|}{97.22$\pm$0.00} & \multicolumn{1}{c|}{0.00$\pm$0.00}  & \multicolumn{1}{c|}{0$\pm$0.00}     & \textbf{36.89$\pm$0.00}       \\
\multicolumn{1}{l|}{}       & MIA               & \multicolumn{1}{c|}{92.40$\pm$0.00} & \multicolumn{1}{c|}{0.32$\pm$0.00}                             & \multicolumn{1}{c|}{0$\pm$0.00}         & 0.20$\pm$0.00                           & \multicolumn{1}{c|}{95.60$\pm$0.00} & \multicolumn{1}{c|}{2.27$\pm$0.01}  & \multicolumn{1}{c|}{1.4$\pm$0.00}   & 0.40$\pm$0.00        \\ \hline
\multicolumn{1}{l|}{MR}     & $\mathcal{D}_{r}$ & \multicolumn{1}{c|}{76.28$\pm$0.00} & \multicolumn{1}{c|}{72.90$\pm$0.45}                            & \multicolumn{1}{c|}{76.12$\pm$0.00}     & 75.59$\pm$0.00                          & \multicolumn{1}{c|}{88.87$\pm$0.00} & \multicolumn{1}{c|}{90.02$\pm$0.22} & \multicolumn{1}{c|}{88.07$\pm$0.00} & 88.82$\pm$0.00       \\
\multicolumn{1}{l|}{}       & $\mathcal{D}_{f}$ & \multicolumn{1}{c|}{80.12$\pm$0.00} & \multicolumn{1}{c|}{0.00$\pm$0.00}                             & \multicolumn{1}{c|}{0$\pm$0.00}         & 0.00$\pm$0.00                           & \multicolumn{1}{c|}{94.88$\pm$0.00} & \multicolumn{1}{c|}{0.00$\pm$0.00}  & \multicolumn{1}{c|}{0$\pm$0.00}     & 0.00$\pm$0.00        \\
\multicolumn{1}{l|}{}       & MIA               & \multicolumn{1}{c|}{95.20$\pm$0.00} & \multicolumn{1}{c|}{0.22$\pm$0.00}                             & \multicolumn{1}{c|}{0.6$\pm$0.00}       & 0.20$\pm$0.00                           & \multicolumn{1}{c|}{92.80$\pm$0.00} & \multicolumn{1}{c|}{0.70$\pm$0.00}  & \multicolumn{1}{c|}{2.2$\pm$0.00}   & 3.80$\pm$0.00        \\ \hline
\multicolumn{1}{l|}{Rkt}    & $\mathcal{D}_{r}$ & \multicolumn{1}{c|}{76.27$\pm$0.00} & \multicolumn{1}{c|}{72.83$\pm$0.42}                            & \multicolumn{1}{c|}{75.12$\pm$0.00}     & 74.54$\pm$0.00                          & \multicolumn{1}{c|}{88.88$\pm$0.00} & \multicolumn{1}{c|}{90.07$\pm$0.09} & \multicolumn{1}{c|}{88.39$\pm$0.00} & 88.90$\pm$0.00       \\
\multicolumn{1}{l|}{}       & $\mathcal{D}_{f}$ & \multicolumn{1}{c|}{80.90$\pm$0.00} & \multicolumn{1}{c|}{0.00$\pm$0.00}                             & \multicolumn{1}{c|}{0$\pm$0.00}         & 0.00$\pm$0.00                           & \multicolumn{1}{c|}{94.70$\pm$0.00} & \multicolumn{1}{c|}{0.00$\pm$0.00}  & \multicolumn{1}{c|}{0$\pm$0.00}     & 0.00$\pm$0.00        \\
\multicolumn{1}{l|}{}       & MIA               & \multicolumn{1}{c|}{93.40$\pm$0.00} & \multicolumn{1}{c|}{1.04$\pm$0.00}                             & \multicolumn{1}{c|}{0$\pm$0.00}         & 2.20$\pm$0.00                           & \multicolumn{1}{c|}{94.40$\pm$0.00} & \multicolumn{1}{c|}{3.23$\pm$0.00}  & \multicolumn{1}{c|}{1.4$\pm$0.00}   & 1.80$\pm$0.00        \\ \hline
\multicolumn{1}{l|}{sea}    & $\mathcal{D}_{r}$ & \multicolumn{1}{c|}{76.23$\pm$0.00} & \multicolumn{1}{c|}{72.83$\pm$0.54}                            & \multicolumn{1}{c|}{75.55$\pm$0.00}     & 73.56$\pm$0.00                          & \multicolumn{1}{c|}{88.91$\pm$0.00} & \multicolumn{1}{c|}{90.27$\pm$0.21} & \multicolumn{1}{c|}{86.98$\pm$0.00} & 87.95$\pm$0.00       \\
\multicolumn{1}{l|}{}       & $\mathcal{D}_{f}$ & \multicolumn{1}{c|}{85.85$\pm$0.00} & \multicolumn{1}{c|}{0.00$\pm$0.00}                             & \multicolumn{1}{c|}{0$\pm$0.00}         & 0.00$\pm$0.00                           & \multicolumn{1}{c|}{90.54$\pm$0.00} & \multicolumn{1}{c|}{0.00$\pm$0.00}  & \multicolumn{1}{c|}{0$\pm$0.00}     & 0.00$\pm$0.00        \\
\multicolumn{1}{l|}{}       & MIA               & \multicolumn{1}{c|}{93.40$\pm$0.00} & \multicolumn{1}{c|}{5.84$\pm$0.02}                             & \multicolumn{1}{c|}{0.2$\pm$0.00}       & 0.60$\pm$0.00                           & \multicolumn{1}{c|}{80.40$\pm$0.00} & \multicolumn{1}{c|}{8.43$\pm$0.02}  & \multicolumn{1}{c|}{4.4$\pm$0.00}   & 3.20$\pm$0.00        \\ \hline
\multicolumn{1}{l|}{ED}     & $\mathcal{D}_{r}$ & \multicolumn{1}{c|}{82.56$\pm$0.00} & \multicolumn{1}{c|}{82.13$\pm$0.23}                            & \multicolumn{1}{c|}{82.92$\pm$0.00}     & 83.15$\pm$0.00                          & \multicolumn{1}{c|}{95.73$\pm$0.00} & \multicolumn{1}{c|}{94.71$\pm$0.14} & \multicolumn{1}{c|}{95.40$\pm$0.00} & 95.82$\pm$0.00       \\
\multicolumn{1}{l|}{}       & $\mathcal{D}_{f}$ & \multicolumn{1}{c|}{82.26$\pm$0.00} & \multicolumn{1}{c|}{0.00$\pm$0.00}                             & \multicolumn{1}{c|}{0$\pm$0.00}         & 1.76$\pm$0.00                           & \multicolumn{1}{c|}{95.03$\pm$0.00} & \multicolumn{1}{c|}{0.00$\pm$0.00}  & \multicolumn{1}{c|}{0$\pm$0.00}     & \textbf{53.53$\pm$0.00}       \\
\multicolumn{1}{l|}{}       & MIA               & \multicolumn{1}{c|}{89.56$\pm$0.00} & \multicolumn{1}{c|}{8.91$\pm$0.01}                             & \multicolumn{1}{c|}{5.60$\pm$0.00}      & 4.16$\pm$0.00                           & \multicolumn{1}{c|}{91.60$\pm$0.00} & \multicolumn{1}{c|}{9.82$\pm$0.01}  & \multicolumn{1}{c|}{5.72$\pm$0.00}  & 1.32$\pm$0.00        \\ \hline
\multicolumn{1}{l|}{NS}     & $\mathcal{D}_{r}$ & \multicolumn{1}{c|}{82.10$\pm$0.00} & \multicolumn{1}{c|}{81.33$\pm$0.22}                            & \multicolumn{1}{c|}{79.32$\pm$0.00}     & 82.33$\pm$0.00                          & \multicolumn{1}{c|}{95.71$\pm$0.00} & \multicolumn{1}{c|}{94.79$\pm$0.11} & \multicolumn{1}{c|}{95.30$\pm$0.00} & 93.63$\pm$0.00       \\
\multicolumn{1}{l|}{}       & $\mathcal{D}_{f}$ & \multicolumn{1}{c|}{91.08$\pm$0.00} & \multicolumn{1}{c|}{0.00$\pm$0.00}                             & \multicolumn{1}{c|}{0$\pm$0.00}         & 0.00$\pm$0.00                           & \multicolumn{1}{c|}{95.37$\pm$0.00} & \multicolumn{1}{c|}{0.00$\pm$0.00}  & \multicolumn{1}{c|}{0$\pm$0.00}     & 0.00$\pm$0.00        \\
\multicolumn{1}{l|}{}       & MIA               & \multicolumn{1}{c|}{88.68$\pm$0.00} & \multicolumn{1}{c|}{3.77$\pm$0.01}                             & \multicolumn{1}{c|}{2.88$\pm$0.00}      & 3.28$\pm$0.00                           & \multicolumn{1}{c|}{85.04$\pm$0.00} & \multicolumn{1}{c|}{4.70$\pm$0.01}  & \multicolumn{1}{c|}{2$\pm$0.00}     & 1.88$\pm$0.00        \\ \hline
\multicolumn{1}{l|}{people} & $\mathcal{D}_{r}$ & \multicolumn{1}{c|}{82.11$\pm$0.00} & \multicolumn{1}{c|}{81.20$\pm$0.19}                            & \multicolumn{1}{c|}{82.13$\pm$0.00}     & 82.31$\pm$0.00                          & \multicolumn{1}{c|}{95.54$\pm$0.00} & \multicolumn{1}{c|}{94.54$\pm$0.14} & \multicolumn{1}{c|}{95.11$\pm$0.00} & 95.33$\pm$0.00       \\
\multicolumn{1}{l|}{}       & $\mathcal{D}_{f}$ & \multicolumn{1}{c|}{90.70$\pm$0.00} & \multicolumn{1}{c|}{0.00$\pm$0.00}                             & \multicolumn{1}{c|}{0$\pm$0.00}         & 0.00$\pm$0.00                           & \multicolumn{1}{c|}{98.54$\pm$0.00} & \multicolumn{1}{c|}{0.00$\pm$0.00}  & \multicolumn{1}{c|}{0$\pm$0.00}     & 0.00$\pm$0.00        \\
\multicolumn{1}{l|}{}       & MIA               & \multicolumn{1}{c|}{91.72$\pm$0.00} & \multicolumn{1}{c|}{1.36$\pm$0.00}                             & \multicolumn{1}{c|}{2.24$\pm$0.00}      & 1.12$\pm$0.00                           & \multicolumn{1}{c|}{89.48$\pm$0.00} & \multicolumn{1}{c|}{1.56$\pm$0.00}  & \multicolumn{1}{c|}{1.48$\pm$0.00}  & 1.20$\pm$0.00        \\ \hline
\multicolumn{1}{l|}{veg}    & $\mathcal{D}_{r}$ & \multicolumn{1}{c|}{82.31$\pm$0.00} & \multicolumn{1}{c|}{81.39$\pm$0.21}                            & \multicolumn{1}{c|}{82.20$\pm$0.00}     & 82.38$\pm$0.00                          & \multicolumn{1}{c|}{95.59$\pm$0.00} & \multicolumn{1}{c|}{94.54$\pm$0.21} & \multicolumn{1}{c|}{95.07$\pm$0.00} & 95.71$\pm$0.00       \\
\multicolumn{1}{l|}{}       & $\mathcal{D}_{f}$ & \multicolumn{1}{c|}{86.90$\pm$0.00} & \multicolumn{1}{c|}{0.00$\pm$0.00}                             & \multicolumn{1}{c|}{0$\pm$0.00}         & 0.00$\pm$0.00                           & \multicolumn{1}{c|}{97.57$\pm$0.00} & \multicolumn{1}{c|}{0.00$\pm$0.00}  & \multicolumn{1}{c|}{0$\pm$0.00}     & 0.00$\pm$0.00        \\
\multicolumn{1}{l|}{}       & MIA               & \multicolumn{1}{c|}{89.52$\pm$0.00} & \multicolumn{1}{c|}{9.74$\pm$0.01}                             & \multicolumn{1}{c|}{13.52$\pm$0.00}     & 16.96$\pm$0.00                          & \multicolumn{1}{c|}{91.32$\pm$0.00} & \multicolumn{1}{c|}{4.41$\pm$0.01}  & \multicolumn{1}{c|}{2.96$\pm$0.00}  & 1.88$\pm$0.00        \\ \hline
\multicolumn{1}{l|}{Veh2}   & $\mathcal{D}_{r}$ & \multicolumn{1}{c|}{82.69$\pm$0.00} & \multicolumn{1}{c|}{82.11$\pm$0.19}                            & \multicolumn{1}{c|}{82.56$\pm$0.00}     & 82.97$\pm$0.00                          & \multicolumn{1}{c|}{95.73$\pm$0.00} & \multicolumn{1}{c|}{94.85$\pm$0.13} & \multicolumn{1}{c|}{93.93$\pm$0.00} & 93.12$\pm$0.00       \\
\multicolumn{1}{l|}{}       & $\mathcal{D}_{f}$ & \multicolumn{1}{c|}{80.41$\pm$0.00} & \multicolumn{1}{c|}{0.00$\pm$0.00}                             & \multicolumn{1}{c|}{0$\pm$0.00}         & 0.00$\pm$0.00                           & \multicolumn{1}{c|}{95.22$\pm$0.00} & \multicolumn{1}{c|}{0.00$\pm$0.00}  & \multicolumn{1}{c|}{0$\pm$0.00}     & 0.00$\pm$0.00        \\
\multicolumn{1}{l|}{}       & MIA               & \multicolumn{1}{c|}{82.56$\pm$0.00} & \multicolumn{1}{c|}{13.54$\pm$0.01}                            & \multicolumn{1}{c|}{5.16$\pm$0.00}      & 6.68$\pm$0.00                           & \multicolumn{1}{c|}{84.04$\pm$0.00} & \multicolumn{1}{c|}{22.96$\pm$0.03} & \multicolumn{1}{c|}{6.04$\pm$0.00}  & 7.04$\pm$0.00        \\ \hline
\multicolumn{1}{l|}{Faces}  & $\mathcal{D}_{r}$ & \multicolumn{1}{c|}{98.52$\pm$0.02} & \multicolumn{1}{c|}{100.00$\pm$0.00} & \multicolumn{1}{c|}{98.16$\pm$0.25} & 98.42$\pm$0.13                          &                                     &                                     &                            & \multicolumn{1}{l}{} \\
\multicolumn{1}{l|}{}       & $\mathcal{D}_{f}$ & \multicolumn{1}{c|}{97.84$\pm$1.99} & \multicolumn{1}{c|}{0.00$\pm$0.00} & \multicolumn{1}{c|}{0.00$\pm$0.00} & 0.00$\pm$0.00 &                                     &                                     &                            & \multicolumn{1}{l}{} \\
\multicolumn{1}{l|}{}       & MIA               & \multicolumn{1}{c|}{34.38$\pm$0.23} & \multicolumn{1}{c|}{0.00$\pm$0.00}  & \multicolumn{1}{c|}{0.74$\pm$1.01} & 1.11$\pm$0.01                           &                                     &                                     &                            & \multicolumn{1}{l}{} \\ \cline{1-6}
\end{tabular}
\caption{Fullclass unlearning on Cifar100 (baby-sea), Cifar20 (ED-Veh2), and Face unlearning. Bold highlights instances where ASSD and SSD differ significantly. While SSD generally performs slightly better due to hyperparameter search for the ResNet18 and Vision Transformer tasks, ASSD is more reliable and does not experience the highlighted performance outliers of SSD.}
\label{tab:fullclass}
\end{table}

\subsection{Unlearning benchmarks}

In the (i) fullclass unlearning tasks shown in Table \ref{tab:fullclass}, SSD and ASSD perform similarly. One difference is the slightly better accuracy of SSD on $D_r$ which is an expected result of parameters chosen for exactly these benchmarks. ASSD, which is parameter-free and automatically chooses its parameters, does not exhibit the performance drops of SSD on various subtasks (highlighted bold in Table \ref{tab:fullclass}). A minor drop in accuracy for increased reliability is a desirable outcome.

In the (ii) subclass unlearning task shown in Table \ref{tab:subclass}, results are similar to the fullclass task with the main difference being that ASSD performs stronger unlearning on the sea subclass.

In the (iii) random unlearning task ASSD and SSD once again perform similarly. ASSD achieves higher accuracy on $D_r$ and a closer match on average MIA to the retrained model. It is important to consider that in a random unlearning scenario, similar images can lead to correct classification of forgotten images due to their similarity (e.g., forgetting one rocket but still having other rockets in the data). Thus, MIA remains high even on a retrained model that has never seen $D_r$

ASSD shares the limitation of SSD in regards to no mathematical guarantees or certifications to unlearn which extends to the automatic selection of $\alpha$ that is based on the assumptions outlined in \ref{sec:method}.

Across all three tasks, ASSD performs similarly to the state-of-the-art method SSD while removing the need for parameter search to find a good balance of unlearning and retained model performance.

\begin{table}[]
\centering
\fontsize{8pt}{10pt}\selectfont
\setlength{\tabcolsep}{1.7pt}
\begin{tabular}{ll|cccc|cccc}
                           &                   & \multicolumn{4}{c|}{ResNet18}                                                                                          & \multicolumn{4}{c}{Vision Transformer}                                                                                 \\ \hline
\multicolumn{1}{l|}{Class} & metric            & \multicolumn{1}{c|}{baseline}       & \multicolumn{1}{c|}{retrain}        & \multicolumn{1}{c|}{ASSD} & SSD            & \multicolumn{1}{c|}{baseline}       & \multicolumn{1}{c|}{retrain}        & \multicolumn{1}{c|}{ASSD} & SSD            \\ \hline
\multicolumn{1}{l|}{baby}  & $\mathcal{D}_{r}$ & \multicolumn{1}{c|}{82.45$\pm$0.00} & \multicolumn{1}{c|}{81.45$\pm$0.3}  & \multicolumn{1}{c|}{78.84$\pm$0.00}     & 79.25$\pm$0.00 & \multicolumn{1}{c|}{95.69$\pm$0.00} & \multicolumn{1}{c|}{94.50$\pm$0.19} & \multicolumn{1}{c|}{94.21$\pm$0.00}     & 95.54$\pm$0.00 \\
\multicolumn{1}{l|}{}      & $\mathcal{D}_{f}$ & \multicolumn{1}{c|}{86.98$\pm$0.00} & \multicolumn{1}{c|}{80.03$\pm$3.4}  & \multicolumn{1}{c|}{0.00$\pm$0.00}     & 8.85$\pm$0.00  & \multicolumn{1}{c|}{96.44$\pm$0.00} & \multicolumn{1}{c|}{93.23$\pm$1.09} & \multicolumn{1}{c|}{40.71$\pm$0.00}     & 94.10$\pm$0.00 \\
\multicolumn{1}{l|}{}      & MIA               & \multicolumn{1}{c|}{90.40$\pm$0.00} & \multicolumn{1}{c|}{44.82$\pm$0.02} & \multicolumn{1}{c|}{2.60$\pm$0.00}     & 1.40$\pm$0.00  & \multicolumn{1}{c|}{91.60$\pm$0.00} & \multicolumn{1}{c|}{77.37$\pm$0.03} & \multicolumn{1}{c|}{80.00$\pm$0.00}     & 77.20$\pm$0.00 \\ \hline
\multicolumn{1}{l|}{lamp}  & $\mathcal{D}_{r}$ & \multicolumn{1}{c|}{82.61$\pm$0.00} & \multicolumn{1}{c|}{81.81$\pm$0.19} & \multicolumn{1}{c|}{79.69$\pm$0.00}     & 79.58$\pm$0.00 & \multicolumn{1}{c|}{95.77$\pm$0.00} & \multicolumn{1}{c|}{94.69$\pm$0.13} & \multicolumn{1}{c|}{95.00$\pm$0.00}     & 95.54$\pm$0.00 \\
\multicolumn{1}{l|}{}      & $\mathcal{D}_{f}$ & \multicolumn{1}{c|}{72.66$\pm$0.00} & \multicolumn{1}{c|}{22.95$\pm$4.72} & \multicolumn{1}{c|}{0.00$\pm$0.00}     & 0.00$\pm$0.00  & \multicolumn{1}{c|}{89.58$\pm$0.00} & \multicolumn{1}{c|}{34.55$\pm$8.62} & \multicolumn{1}{c|}{3.56$\pm$0.00}     & 14.58$\pm$0.00 \\
\multicolumn{1}{l|}{}      & MIA               & \multicolumn{1}{c|}{83.40$\pm$0.00} & \multicolumn{1}{c|}{4.29$\pm$0.01}  & \multicolumn{1}{c|}{1.40$\pm$0.00}     & 2.00$\pm$0.00  & \multicolumn{1}{c|}{81.00$\pm$0.00} & \multicolumn{1}{c|}{5.60$\pm$0.02}  & \multicolumn{1}{c|}{5.60$\pm$0.00}     & 3.2$\pm$0.00   \\ \hline
\multicolumn{1}{l|}{MR}    & $\mathcal{D}_{r}$ & \multicolumn{1}{c|}{82.60$\pm$0.00} & \multicolumn{1}{c|}{81.66$\pm$0.21} & \multicolumn{1}{c|}{80.40$\pm$0.00}     & 80.05$\pm$0.00 & \multicolumn{1}{c|}{95.69$\pm$0.00} & \multicolumn{1}{c|}{94.60$\pm$0.13} & \multicolumn{1}{c|}{92.13$\pm$0.00}     & 95.51$\pm$0.00 \\
\multicolumn{1}{l|}{}      & $\mathcal{D}_{f}$ & \multicolumn{1}{c|}{72.22$\pm$0.00} & \multicolumn{1}{c|}{10.48$\pm$2.02} & \multicolumn{1}{c|}{0.78$\pm$0.00}     & 0.78$\pm$0.00  & \multicolumn{1}{c|}{97.05$\pm$0.00} & \multicolumn{1}{c|}{26.57$\pm$6.41} & \multicolumn{1}{c|}{0.00$\pm$0.00}     & 6.68$\pm$0.00  \\
\multicolumn{1}{l|}{}      & MIA               & \multicolumn{1}{c|}{86.80$\pm$0.00} & \multicolumn{1}{c|}{2.27$\pm$0.00}  & \multicolumn{1}{c|}{4.40$\pm$0.00}     & 4.00$\pm$0.00  & \multicolumn{1}{c|}{77.80$\pm$0.00} & \multicolumn{1}{c|}{2.34$\pm$0.01}  & \multicolumn{1}{c|}{0.60$\pm$0.00}     & 0.40$\pm$0.00  \\ \hline
\multicolumn{1}{l|}{Rkt}   & $\mathcal{D}_{r}$ & \multicolumn{1}{c|}{82.54$\pm$0.00} & \multicolumn{1}{c|}{81.54$\pm$0.24} & \multicolumn{1}{c|}{82.37$\pm$0.00}     & 82.43$\pm$0.00 & \multicolumn{1}{c|}{95.73$\pm$0.00} & \multicolumn{1}{c|}{94.61$\pm$0.13} & \multicolumn{1}{c|}{95.22$\pm$0.00}     & 95.13$\pm$0.00 \\
\multicolumn{1}{l|}{}      & $\mathcal{D}_{f}$ & \multicolumn{1}{c|}{79.34$\pm$0.00} & \multicolumn{1}{c|}{10.74$\pm$3.4}  & \multicolumn{1}{c|}{2.17$\pm$0.00}     & 2.17$\pm$0.00  & \multicolumn{1}{c|}{94.53$\pm$0.00} & \multicolumn{1}{c|}{22.26$\pm$8.34} & \multicolumn{1}{c|}{5.90$\pm$0.00}     & 5.12$\pm$0.00  \\
\multicolumn{1}{l|}{}      & MIA               & \multicolumn{1}{c|}{89.40$\pm$0.00} & \multicolumn{1}{c|}{3.85$\pm$0.01}  & \multicolumn{1}{c|}{9.40$\pm$0.00}     & 10.80$\pm$0.00 & \multicolumn{1}{c|}{80.40$\pm$0.00} & \multicolumn{1}{c|}{3.44$\pm$0.01}  & \multicolumn{1}{c|}{3.80$\pm$0.00}     & 5.40$\pm$0.00  \\ \hline
\multicolumn{1}{l|}{sea}   & $\mathcal{D}_{r}$ & \multicolumn{1}{c|}{82.37$\pm$0.00} & \multicolumn{1}{c|}{81.30$\pm$0.27} & \multicolumn{1}{c|}{80.60$\pm$0.00}     & 81.72$\pm$0.00 & \multicolumn{1}{c|}{95.67$\pm$0.00} & \multicolumn{1}{c|}{94.55$\pm$0.22} & \multicolumn{1}{c|}{94.89$\pm$0.00}     & 95.57$\pm$0.00 \\
\multicolumn{1}{l|}{}      & $\mathcal{D}_{f}$ & \multicolumn{1}{c|}{96.27$\pm$0.00} & \multicolumn{1}{c|}{91.47$\pm$1.92} & \multicolumn{1}{c|}{\textbf{34.20$\pm$0.00}}     & 75.35$\pm$0.00 & \multicolumn{1}{c|}{99.22$\pm$0.00} & \multicolumn{1}{c|}{95.12$\pm$0.81} & \multicolumn{1}{c|}{84.81$\pm$0.00}     & 97.05$\pm$0.00 \\
\multicolumn{1}{l|}{}      & MIA               & \multicolumn{1}{c|}{90.80$\pm$0.00} & \multicolumn{1}{c|}{52.09$\pm$0.03} & \multicolumn{1}{c|}{\textbf{2.60$\pm$0.00}}     & 21.80$\pm$0.00 & \multicolumn{1}{c|}{88.40$\pm$0.00} & \multicolumn{1}{c|}{65.96$\pm$0.04} & \multicolumn{1}{c|}{\textbf{38.20$\pm$0.00}}     & 82.20$\pm$0.00 \\ \hline
\end{tabular}
\caption{Subclass unlearning on Cifar20. Bold highlights instances where ASSD and SSD differ significantly. ASSD performs significantly stronger unlearning on the sea (ViT) instances.}
\label{tab:subclass}
\end{table}

\begin{table}[]
\centering
\fontsize{8pt}{10pt}\selectfont
\setlength{\tabcolsep}{1.7pt}
\begin{tabular}{l|lccc|lccc}
                  & \multicolumn{4}{c|}{ResNet18}                                                                                                                                              & \multicolumn{4}{c}{Vision Transformer}                                                                                                                                                               \\ \hline
metric            & \multicolumn{1}{c|}{baseline}       & \multicolumn{1}{c|}{retrain}                                  & \multicolumn{1}{c|}{ASSD} & SSD                                      & \multicolumn{1}{c|}{baseline}                                 & \multicolumn{1}{c|}{retrain}                                  & \multicolumn{1}{c|}{ASSD} & SSD                                      \\ \hline
$\mathcal{D}_{r}$ & \multicolumn{1}{l|}{90.71$\pm$0.00} & \multicolumn{1}{c|}{91.45$\pm$0.11} & \multicolumn{1}{c|}{90.67$\pm$0.07}     & 88.68$\pm$3.36                           & \multicolumn{1}{l|}{98.88$\pm$0.00} & \multicolumn{1}{c|}{98.61$\pm$0.08}                           & \multicolumn{1}{c|}{98.85$\pm$0.04}     & 98.01$\pm$1.56 \\
$\mathcal{D}_{f}$ & \multicolumn{1}{l|}{95.30$\pm$2.08} & \multicolumn{1}{c|}{94.10$\pm$2.00} & \multicolumn{1}{c|}{95.30$\pm$2.16}     & 93.61$\pm$4.99 & \multicolumn{1}{l|}{100.00$\pm$0.00}                          & \multicolumn{1}{c|}{98.80$\pm$0.76} & \multicolumn{1}{c|}{99.64$\pm$0.58}     & 98.07$\pm$2.35 \\
MIA               & \multicolumn{1}{l|}{75.78$\pm$0.04} & \multicolumn{1}{c|}{74.22$\pm$0.04} & \multicolumn{1}{c|}{75.54$\pm$4.17}     & 72.65$\pm$0.05 & \multicolumn{1}{l|}{90.76$\pm$0.03}                           & \multicolumn{1}{c|}{91.77$\pm$0.02} & \multicolumn{1}{c|}{92.89$\pm$4.11}     & 85.54$\pm$0.11 \\ \hline
\end{tabular}
\caption{Random unlearning on Cifar 10. Due to a large number of similar images in the retain data out of which we forget random observations, values hardly change as generalization covers nearly all forgotten cases (e.g., retrain MIA for ViT even increases).}
\label{tab:random}
\end{table}

\subsection{Model improvement via error unlearning}

\begin{figure} []
    \centering
    \includegraphics*[width=1.0\columnwidth]{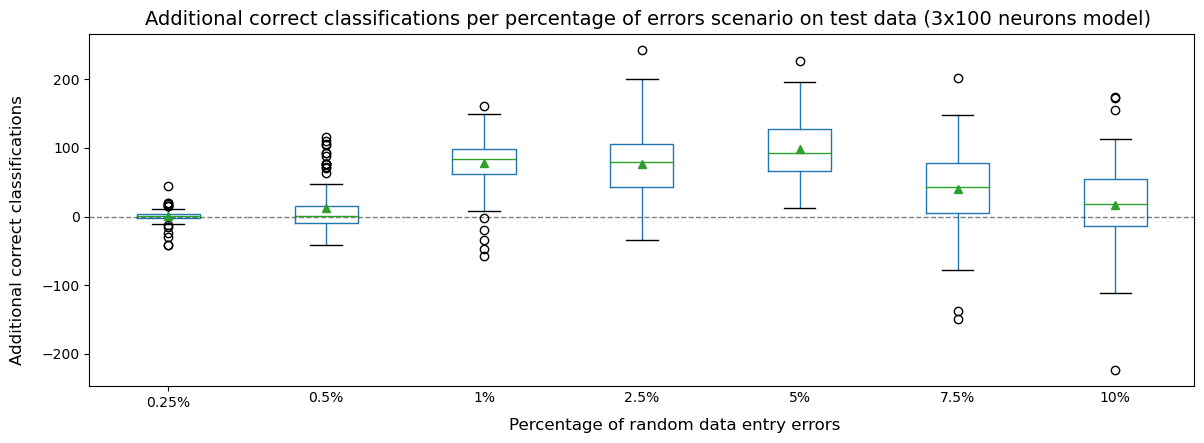}
    \caption{Model accuracy improvement due to unlearning compared with the baseline model trained on data with random errors for 100 random scenarios using a neural network with three layers of 100 neurons each.}
    \label{fig:3x100_boxes}
\end{figure}

In the error unlearning task we compare unlearning using ASSD to fine-tuning of the baseline model and retraining on $D_r$. Table \ref{tab:error_unlearning} shows the accuracy of each model on the train and test data for three different model sizes (shown as\textit{ layers $\cdot$ neurons per layer}) and six different random error amounts ranging from 0.25\% to 10\%.
The best-performing model amongst baseline, fine-tuning and ASSD is highlighted in bold. Retraining is not highlighted, as full retraining is much more resource and time-intensive compared to the other approaches and acts as a reference for what a model trained on $D_r$ would achieve. It is notable, that while retraining outperforms the other methods on low error percentages ($\leq1\%$), fine-tuning and ASSD tend to achieve similar or better results at higher error percentages. A possible explanation for this is that removing a significant percentage of training data from $D$ (e.g., 5\% $D_f$) for retraining hurts performance more than learning the errors in the first place and then unlearning them by fine-tuning or unlearning. The error data in the training might help the model to be more robust and thus allow for better performance combined with fine-tuning or unlearning.
As we start to go beyond 5\% errors in the training data, the performance of ASSD starts to deteriorate with significant dips in accuracy. This is expected as ASSD performs unlearning without any kind of repair steps (i.e., additional training epochs to increase accuracy after unlearning). With error rates of 7.5\% and 10\% the amount of changes needed in the model to remove this information becomes significant and accuracy dips occur. Such high error rates likely indicate another systematic problem in data collection or preparation that should be corrected instead.
Fig. \ref{fig:3x100_boxes} shows the performance gain as additional correct classifications on the test data. While unlearning still presents a net gain on 10\% error data in Fig. \ref{fig:3x100_boxes}, a clear downward trend after 5\% can be observed due to the mentioned large changes to the model without a repair step.

\begin{figure}[h!]
    \centering
    \begin{subfigure}{\textwidth}
        \centering
        \includegraphics[width=1\textwidth]{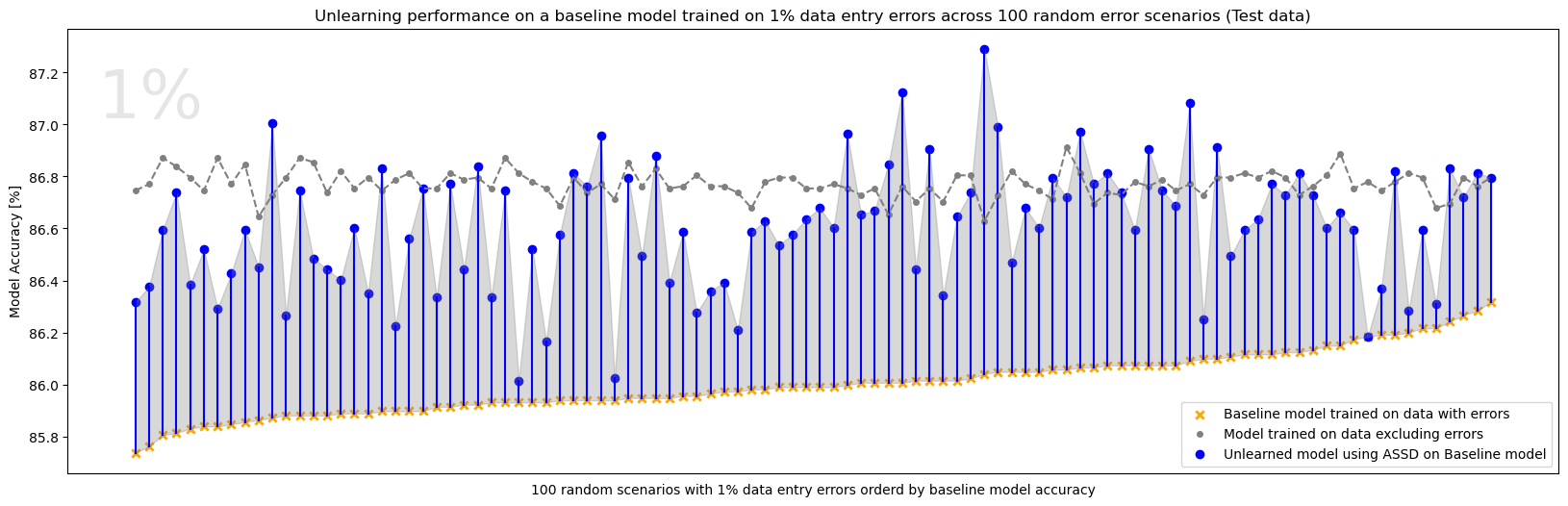}
        \caption{Large model with 5x250 layers and neurons.}
        \label{fig:1perc_detail}
    \end{subfigure}
    
    \vspace{0.5cm} 
    
    \begin{subfigure}{\textwidth}
        \centering
        \includegraphics[width=1\textwidth]{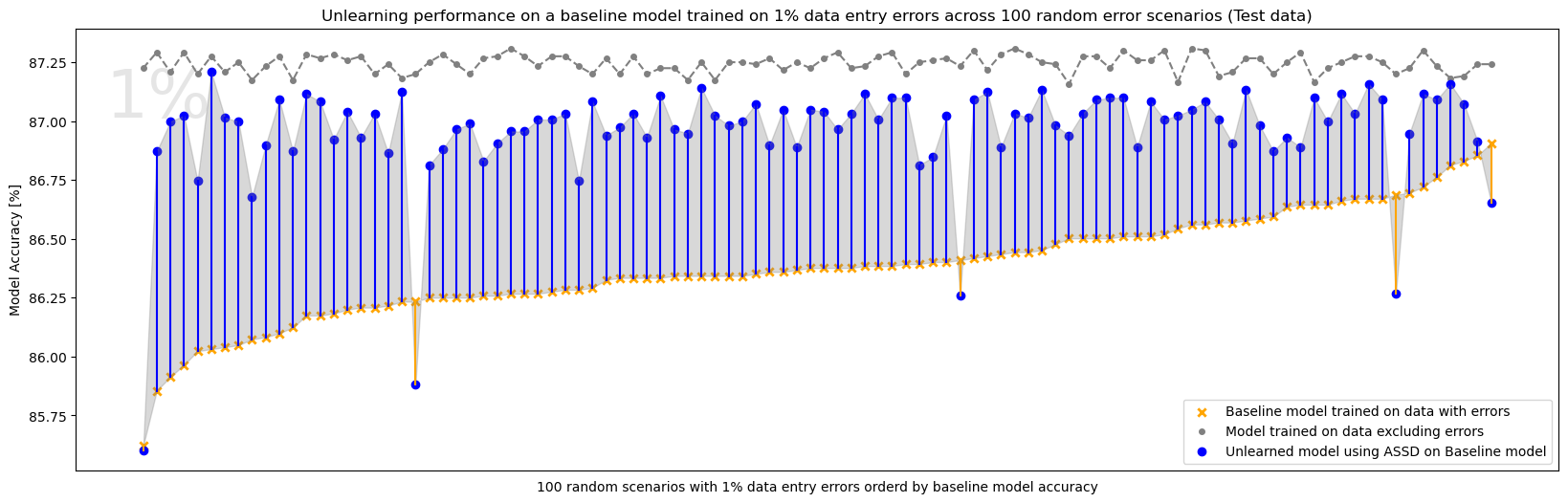}
        \caption{Medium model with 3x100 layers and neurons.}
        \label{fig:2perc_detail}
    \end{subfigure}
    
    \vspace{0.5cm} 
    
    \begin{subfigure}{\textwidth}
        \centering
        \includegraphics[width=1\textwidth]{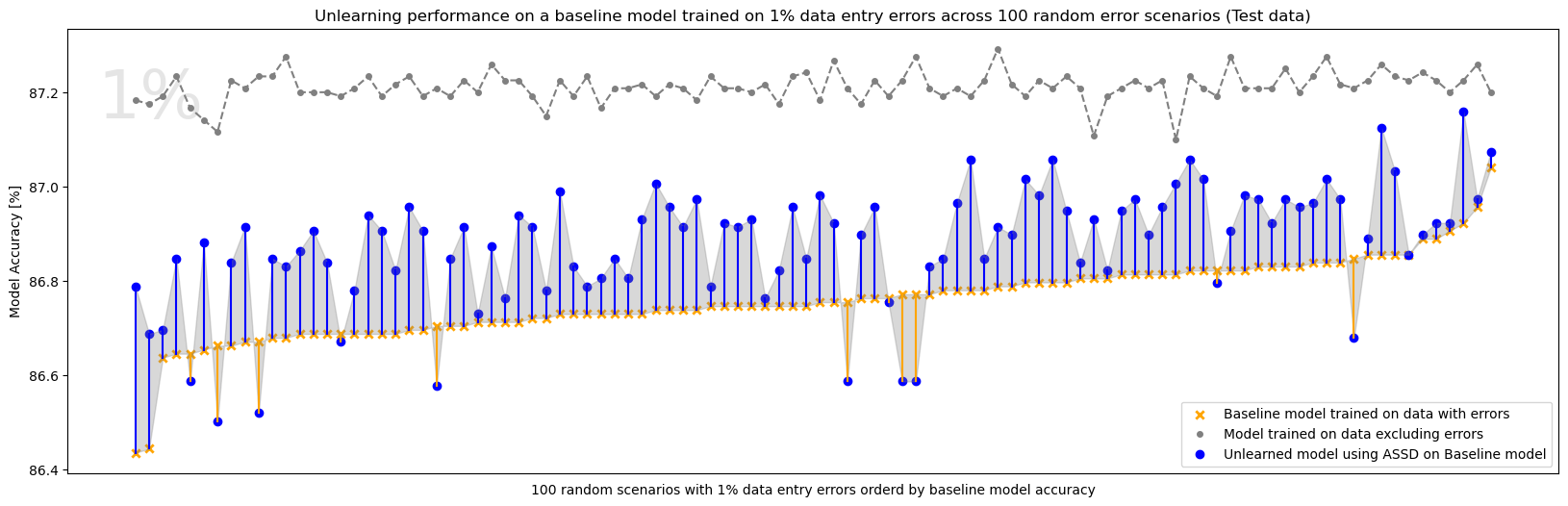}
        \caption{Small model with 2x50 layers and neurons.}
        \label{fig:3perc_detail}
    \end{subfigure}
    
    \caption{Model accuracy improvement due to unlearning compared with the baseline model trained on data with 1\% random errors and a retrained model on data excluding the errors. 100 error scenarios with randomly chosen observations to be mislabeled are shown ordered by the accuracy of the baseline model on the test data.}
    \label{fig:overall_figure}
\end{figure}

Fig. \ref{fig:1perc_detail}, \ref{fig:2perc_detail}, and \ref{fig:3perc_detail} shows a detailed view of the unlearning improvement on the test data in the 1\% label error scenario for all three model sizes. The baseline model accuracies for 100 random scenarios with 1\% mislabeled data each are plotted in increasing order. We can observe that unlearning increases the performance on the large model more than on the small model. A possible hypothesis for this is that larger (overparameterized) models have a greater ability to memorize wrong data compared to a smaller model that is forced to generalize more due to its limited parameter count. This observation is in line with the findings of \citet{shortlongtail} on memorization of information in deep learning models. 

Overall, ASSD reliably increases model performance by unlearning small errors on the shown test scenarios and outperforms fine-tuning.

\begin{table}[]
\centering
\fontsize{8pt}{12pt}\selectfont
\setlength{\tabcolsep}{2.5pt}
\begin{tabular}{c|c|c|c|c|c|c|c}
\cline{1-8}
Model size & Data             & Error              & Baseline Accuracy & Retrain Accuracy & Finetune Accuracy & Unlearn Accuracy & p-value* \\ \hline
5x250      & \multirow{18}{*}{Test}  & \multirow{3}{*}{0.25\%} & 85.95 ± 0.06      & 86.19 ± 0.10     & 85.94 ± 0.07      & \textbf{85.98 ± 0.09}     & \textless{}0.05             \\ \cline{4-8} 
3x100      &                         &                       & 86.34 ± 0.18      & 86.62 ± 0.14     & 86.32 ± 0.13      & \textbf{86.35 ± 0.18}     & 0.92                        \\ \cline{4-8} 
2x50       &                         &                       & 86.72 ± 0.03      & 86.96 ± 0.03     & 86.72 ± 0.03      & \textbf{86.76 ± 0.04}     & \textless{}0.05             \\ \cline{1-1} \cline{3-8} 
5x250      &                         & \multirow{3}{*}{1.0\%}  & 86.00 ± 0.12      & 86.77 ± 0.05     & 86.21 ± 0.14      & \textbf{86.61 ± 0.24}     & \textless{}0.05             \\ \cline{4-8} 
3x100      &                         &                       & 86.38 ± 0.22      & 87.24 ± 0.04     & 86.87 ± 0.16      & \textbf{86.96 ± 0.23}     & \textless{}0.05             \\ \cline{4-8} 
2x50       &                         &                       & 86.76 ± 0.09      & 87.21 ± 0.03     & 86.85 ± 0.08      & \textbf{86.88 ± 0.13}     & \textless{}0.05             \\ \cline{1-1} \cline{3-8} 
5x250      &                         & \multirow{3}{*}{2.5\%}  & 86.18 ± 0.21      & 86.48 ± 0.11     & 86.11 ± 0.18      & \textbf{87.10 ± 0.25}     & \textless{}0.05             \\ \cline{4-8} 
3x100      &                         &                       & 86.54 ± 0.25      & 86.76 ± 0.07     & 86.48 ± 0.23      & \textbf{87.10 ± 0.25}     & \textless{}0.05             \\ \cline{4-8} 
2x50       &                         &                       & 86.92 ± 0.11      & 87.02 ± 0.05     & 86.89 ± 0.11      & \textbf{87.05 ± 0.16}     & \textless{}0.05             \\ \cline{1-1} \cline{3-8} 
5x250      &                         & \multirow{3}{*}{5.0\%}  & 86.39 ± 0.24      & 86.64 ± 0.08     & 86.34 ± 0.20      & \textbf{86.86 ± 0.46}     & \textless{}0.05             \\ \cline{4-8} 
3x100      &                         &                       & 86.70 ± 0.30      & 86.69 ± 0.09     & 86.68 ± 0.24      & \textbf{87.42 ± 0.17}     & \textless{}0.05             \\ \cline{4-8} 
2x50       &                         &                       & 87.07 ± 0.11      & 87.24 ± 0.07     & 87.06 ± 0.11      & \textbf{87.47 ± 0.19}     & \textless{}0.05             \\ \cline{1-1} \cline{3-8} 
5x250      &                         & \multirow{3}{*}{7.5\%}  & 86.51 ± 0.23      & 86.85 ± 0.08     & \textbf{86.60 ± 0.20}      & 85.36 ± 1.38     & \textless{}0.05             \\ \cline{4-8} 
3x100      &                         &                       & 86.84 ± 0.26      & 86.40 ± 0.21     & 86.53 ± 0.33      & \textbf{87.14 ± 0.36}     & \textless{}0.05             \\ \cline{4-8} 
2x50       &                         &                       & \textbf{87.16 ± 0.11 }     & 86.79 ± 0.10     & 87.02 ± 0.11      & 87.10 ± 0.72  & 0.46         

     \\ \cline{1-1} \cline{3-8} 
5x250      &                         & \multirow{3}{*}{10\%}  & \textbf{86.56 ± 0.28}      & 86.73 ± 0.11     & 86.54 ± 0.25    & 82.53 ± 2.03     & \textless{}0.05             \\ \cline{4-8} 
3x100      &                         &                       & 86.92 ± 0.31      & 86.62 ± 0.21     & 86.57 ± 0.36      & \textbf{87.04 ± 0.40}     & \textless{}0.05             \\ \cline{4-8} 
2x50       &                         &                       & \textbf{87.25 ± 0.12 }     & 87.01 ± 0.10     & 87.06 ± 0.12      & 87.13 ± 0.57  & \textless{}0.05    
\\ \hline \hline
5x250      & \multirow{18}{*}{Train} & \multirow{3}{*}{0.25\%} & 86.49 ± 0.05      & 86.62 ± 0.05     & 86.47 ± 0.05      & \textbf{86.52 ± 0.07}     & \textless{}0.05             \\ \cline{4-8} 
3x100      &                         &                       & 87.23 ± 0.05      & 87.38 ± 0.06     & \textbf{87.25 ± 0.04}      & 87.24 ± 0.05     & 0.13                        \\ \cline{4-8} 
2x50       &                         &                       & 87.08 ± 0.02      & 87.19 ± 0.02     & 87.09 ± 0.03      & \textbf{87.10 ± 0.04}     & \textless{}0.05             \\ \cline{1-1} \cline{3-8} 
5x250      &                         & \multirow{3}{*}{1.0\%}  & 86.50 ± 0.07      & 87.07 ± 0.05     & 86.72 ± 0.08      & \textbf{86.94 ± 0.14}     & \textless{}0.05             \\ \cline{4-8} 
3x100      &                         &                       & 87.23 ± 0.07      & 87.57 ± 0.05     & \textbf{87.57 ± 0.07}      & 87.35 ± 0.08     & \textless{}0.05             \\ \cline{4-8} 
2x50       &                         &                       & 87.16 ± 0.06      & 87.54 ± 0.03     & \textbf{87.27 ± 0.05}      & 87.21 ± 0.10     & \textless{}0.05             \\ \cline{1-1} \cline{3-8} 
5x250      &                         & \multirow{3}{*}{2.5\%}  & 86.62 ± 0.10      & 86.86 ± 0.06     & 86.57 ± 0.09      & \textbf{87.48 ± 0.20}     & \textless{}0.05             \\ \cline{4-8} 
3x100      &                         &                       & 87.28 ± 0.08      & 87.16 ± 0.06     & 87.25 ± 0.08      & \textbf{87.53 ± 0.12}     & \textless{}0.05             \\ \cline{4-8} 
2x50       &                         &                       & 87.31 ± 0.08      & 87.35 ± 0.04     & 87.28 ± 0.08      & \textbf{87.44 ± 0.12}     & \textless{}0.05             \\ \cline{1-1} \cline{3-8} 
5x250      &                         & \multirow{3}{*}{5.0\%}  & 86.79 ± 0.14      & 87.06 ± 0.10     & 86.76 ± 0.13      & \textbf{87.28 ± 0.37}     & \textless{}0.05             \\ \cline{4-8} 
3x100      &                         &                       & 87.35 ± 0.10      & 87.19 ± 0.08     & 87.33 ± 0.10      & \textbf{87.71 ± 0.18}     & \textless{}0.05             \\ \cline{4-8} 
2x50       &                         &                       & 87.45 ± 0.08      & 87.51 ± 0.06     & 87.45 ± 0.08      & \textbf{87.82 ± 0.17}     & \textless{}0.05             \\ \cline{1-1} \cline{3-8} 
5x250      &                         & \multirow{3}{*}{7.5\%}  & 86.91 ± 0.14      & 87.21 ± 0.10     & \textbf{86.92 ± 0.12}      & 85.77 ± 1.14     & \textless{}0.05             \\ \cline{4-8} 
3x100      &                         &                       & 87.39 ± 0.12      & 86.95 ± 0.11     & 87.07 ± 0.14      & \textbf{87.41 ± 0.29}     & 0.59                        \\ \cline{4-8} 
2x50       &                         &                       & 87.54 ± 0.10      & 87.26 ± 0.08     & 87.42 ± 0.08      & \textbf{87.56 ± 0.60}     & 0.83   

     \\ \cline{1-1} \cline{3-8} 
5x250      &                         & \multirow{3}{*}{10\%}  & \textbf{87.01 ± 0.17}      & 86.84 ± 0.09     & 86.93 ± 0.15    & 82.90 ± 2.00     & \textless{}0.05             \\ \cline{4-8} 
3x100      &                         &                       & \textbf{87.42 ± 0.12}      & 87.25 ± 0.10     & 87.11 ± 0.14      & 87.24 ± 0.37     & \textless{}0.05             \\ \cline{4-8} 
2x50       &                         &                       & \textbf{87.65 ± 0.11 }     & 87.33 ± 0.09     & 87.47 ± 0.10      & 87.53 ± 0.49  & \textless{}0.05 

\\ \hline
\end{tabular}
\caption{Model unlearning performance for 100 random error scenarios with three different model sizes and 5 different error percentages. Best performing model by mean value (excluding. retraining) in bold. Baseline model trained on data with errors, retain model trained on data excluding the errors, fine-tune accuracy for a fine-tuned model starting from the baseline model, and unlearned model based on the baseline model with ASSD.
*Statistical significance test for Baseline and Unlearning accuracy}
\label{tab:error_unlearning}
\end{table}

\newpage

\section{Conclusion}

We present a novel parameter selection method for SSD to allow for parameter-free unlearning in practice. Our method uses insights on information memorization in overparameterized neural networks to automatically unlearn undesired information while preserving model performance on the remaining data. We evaluate ASSD against SSD on its original benchmarks and show that ASSD achieves similar performance with increased reliability and no need to choose parameters. This enables practitioners to use unlearning without requiring expert knowledge.
We then apply ASSD to error unlearning scenarios, experimentally demonstrating the viability of our method to remove the influence of wrongfully labelled data from an already trained model to increase model performance on future data.

\bibliography{iclr2021_conference}

\begin{thebibliography}{24}
\providecommand{\natexlab}[1]{#1}
\providecommand{\url}[1]{\texttt{#1}}
\expandafter\ifx\csname urlstyle\endcsname\relax
  \providecommand{\doi}[1]{doi: #1}\else
  \providecommand{\doi}{doi: \begingroup \urlstyle{rm}\Url}\fi

\bibitem[Agarap(2018)]{agarap2018deep}
Abien~Fred Agarap.
\newblock Deep learning using rectified linear units (relu).
\newblock \emph{arXiv preprint arXiv:1803.08375}, 2018.

\bibitem[Akiba et~al.(2019)Akiba, Sano, Yanase, Ohta, and Koyama]{akiba2019optuna}
Takuya Akiba, Shotaro Sano, Toshihiko Yanase, Takeru Ohta, and Masanori Koyama.
\newblock Optuna: A next-generation hyperparameter optimization framework.
\newblock In \emph{Proceedings of the 25th ACM SIGKDD international conference on knowledge discovery \& data mining}, pp.\  2623--2631, 2019.

\bibitem[Amari(1993)]{amari1993backpropagation}
Shun-ichi Amari.
\newblock Backpropagation and stochastic gradient descent method.
\newblock \emph{Neurocomputing}, 5\penalty0 (4-5):\penalty0 185--196, 1993.

\bibitem[Burak(2020)]{pinsds}
Burak.
\newblock Pinterest face recognition dataset.
\newblock www.kaggle.com/datasets/hereisburak/pins-face-recognition, 2020.

\bibitem[Cao \& Yang(2015)Cao and Yang]{cao2015towards}
Yinzhi Cao and Junfeng Yang.
\newblock Towards making systems forget with machine unlearning.
\newblock In \emph{2015 IEEE symposium on security and privacy}, pp.\  463--480. IEEE, 2015.

\bibitem[Chundawat et~al.(2023)Chundawat, Tarun, Mandal, and Kankanhalli]{chundawat2023can}
Vikram~S Chundawat, Ayush~K Tarun, Murari Mandal, and Mohan Kankanhalli.
\newblock Can bad teaching induce forgetting? unlearning in deep networks using an incompetent teacher.
\newblock In \emph{Proceedings of the AAAI Conference on Artificial Intelligence}, volume~37, pp.\  7210--7217, 2023.

\bibitem[Constante et~al.(2019)Constante, Silvia, and Pereira]{dataset}
Fabian Constante, Fernando Silvia, and Antonio Pereira.
\newblock Dataco smart supply chain for big data analysis, 2019.

\bibitem[Dosovitskiy et~al.(2020)Dosovitskiy, Beyer, Kolesnikov, Weissenborn, Zhai, Unterthiner, Dehghani, Minderer, Heigold, Gelly, et~al.]{dosovitskiy2020image_vit}
Alexey Dosovitskiy, Lucas Beyer, Alexander Kolesnikov, Dirk Weissenborn, Xiaohua Zhai, Thomas Unterthiner, Mostafa Dehghani, Matthias Minderer, Georg Heigold, Sylvain Gelly, et~al.
\newblock An image is worth 16x16 words: Transformers for image recognition at scale.
\newblock \emph{arXiv preprint arXiv:2010.11929}, 2020.

\bibitem[Feldman(2019)]{shortlongtail}
Vitaly Feldman.
\newblock Does learning require memorization? {A} short tale about a long tail.
\newblock \emph{arXiv preprint arXiv:1906.05271}, 2019.

\bibitem[Foster et~al.(2023)Foster, Schoepf, and Brintrup]{foster2023fast}
Jack Foster, Stefan Schoepf, and Alexandra Brintrup.
\newblock Fast machine unlearning without retraining through selective synaptic dampening.
\newblock \emph{arXiv preprint arXiv:2308.07707}, 2023.

\bibitem[Foster et~al.(2024)Foster, Fogarty, Schoepf, Öztireli, and Brintrup]{foster2024zeroshot}
Jack Foster, Kyle Fogarty, Stefan Schoepf, Cengiz Öztireli, and Alexandra Brintrup.
\newblock Zero-shot machine unlearning at scale via lipschitz regularization.
\newblock \emph{arXiv preprint arXiv:2402.01401}, 2024.

\bibitem[Golatkar et~al.(2020)Golatkar, Achille, and Soatto]{golatkar2020eternal}
Aditya Golatkar, Alessandro Achille, and Stefano Soatto.
\newblock Eternal sunshine of the spotless net: Selective forgetting in deep networks.
\newblock In \emph{Proceedings of the IEEE/CVF Conference on Computer Vision and Pattern Recognition}, pp.\  9304--9312, 2020.

\bibitem[Hartvigsen et~al.(2023)Hartvigsen, Sankaranarayanan, Palangi, Kim, and Ghassemi]{edit_hartvigsen2023aging}
Thomas Hartvigsen, Swami Sankaranarayanan, Hamid Palangi, Yoon Kim, and Marzyeh Ghassemi.
\newblock Aging with grace: Lifelong model editing with discrete key-value adaptors.
\newblock \emph{arXiv preprint arXiv:2211.11031}, 2023.

\bibitem[He et~al.(2016)He, Zhang, Ren, and Sun]{he2016deep}
Kaiming He, Xiangyu Zhang, Shaoqing Ren, and Jian Sun.
\newblock Deep residual learning for image recognition.
\newblock In \emph{Proceedings of the IEEE conference on computer vision and pattern recognition}, pp.\  770--778, 2016.

\bibitem[Ioffe \& Szegedy(2015)Ioffe and Szegedy]{ioffe2015batch}
Sergey Ioffe and Christian Szegedy.
\newblock Batch normalization: Accelerating deep network training by reducing internal covariate shift.
\newblock In \emph{International conference on machine learning}, pp.\  448--456. pmlr, 2015.

\bibitem[Kodge et~al.(2023)Kodge, Saha, and Roy]{kodge2023deep}
Sangamesh Kodge, Gobinda Saha, and Kaushik Roy.
\newblock Deep unlearning: Fast and efficient training-free approach to controlled forgetting.
\newblock \emph{arXiv preprint arXiv:2312.00761}, 2023.

\bibitem[Krizhevsky \& Hinton(2010)Krizhevsky and Hinton]{CIFAR_krizhevsky2010convolutional}
Alex Krizhevsky and Geoff Hinton.
\newblock Convolutional deep belief networks on cifar-10.
\newblock \emph{Unpublished manuscript}, 40\penalty0 (7):\penalty0 1--9, 2010.

\bibitem[Kurmanji et~al.(2023{\natexlab{a}})Kurmanji, Triantafillou, and Triantafillou]{kurmanji2023machine_database}
Meghdad Kurmanji, Eleni Triantafillou, and Peter Triantafillou.
\newblock Machine unlearning in learned databases: An experimental analysis.
\newblock \emph{arXiv preprint arXiv:2311.17276}, 2023{\natexlab{a}}.

\bibitem[Kurmanji et~al.(2023{\natexlab{b}})Kurmanji, Triantafillou, Hayes, and Triantafillou]{kurmanji2023unbounded}
Meghdad Kurmanji, Peter Triantafillou, Jamie Hayes, and Eleni Triantafillou.
\newblock Towards unbounded machine unlearning.
\newblock \emph{arXiv preprint arXiv:2302.09880}, 2023{\natexlab{b}}.

\bibitem[Northcutt et~al.(2021)Northcutt, Athalye, and Mueller]{northcutt2021pervasive}
Curtis~G. Northcutt, Anish Athalye, and Jonas Mueller.
\newblock Pervasive label errors in test sets destabilize machine learning benchmarks.
\newblock \emph{arXiv preprint arXiv:2103.14749}, 2021.

\bibitem[Santurkar et~al.(2021)Santurkar, Tsipras, Elango, Bau, Torralba, and Madry]{edit_santurkar2021editing}
Shibani Santurkar, Dimitris Tsipras, Mahalaxmi Elango, David Bau, Antonio Torralba, and Aleksander Madry.
\newblock Editing a classifier by rewriting its prediction rules.
\newblock \emph{arXiv preprint arXiv:2112.01008}, 2021.

\bibitem[Tanno et~al.(2022)Tanno, F~Pradier, Nori, and Li]{tanno2022repairing}
Ryutaro Tanno, Melanie F~Pradier, Aditya Nori, and Yingzhen Li.
\newblock Repairing neural networks by leaving the right past behind.
\newblock \emph{Advances in Neural Information Processing Systems}, 35:\penalty0 13132--13145, 2022.

\bibitem[Tarun et~al.(2023)Tarun, Chundawat, Mandal, and Kankanhalli]{tarun2023fast}
Ayush~K Tarun, Vikram~S Chundawat, Murari Mandal, and Mohan Kankanhalli.
\newblock Fast yet effective machine unlearning.
\newblock \emph{IEEE Transactions on Neural Networks and Learning Systems}, 2023.

\bibitem[Wang et~al.(2022)Wang, Kale, Nori, Stella, Nunnally, Chau, Vorvoreanu, Wortman~Vaughan, and Caruana]{edit_Wang_2022}
Zijie~J. Wang, Alex Kale, Harsha Nori, Peter Stella, Mark~E. Nunnally, Duen~Horng Chau, Mihaela Vorvoreanu, Jennifer Wortman~Vaughan, and Rich Caruana.
\newblock Interpretability, then what? editing machine learning models to reflect human knowledge and values.
\newblock In \emph{Proceedings of the 28th ACM SIGKDD Conference on Knowledge Discovery and Data Mining}, KDD ’22. ACM, August 2022.
\newblock \doi{10.1145/3534678.3539074}.

\end{thebibliography}
\bibliographystyle{iclr2021_conference}

\appendix

\end{document}